\pdfoutput=1
\documentclass{article}
\usepackage{spconf,amsmath,graphicx,booktabs,array,tabularx,ragged2e}
\newcolumntype{L}{>{\hsize=1.9\hsize\RaggedRight\arraybackslash}X}
\newcolumntype{R}{>{\hsize=0.8875\hsize\raggedleft\arraybackslash}X}
\usepackage{times}
\usepackage{enumitem}
\usepackage{amssymb}
\usepackage{url}
\usepackage{multirow}
\usepackage{graphicx}
\usepackage{pgfplots}
\pgfplotsset{compat=1.18}
\usetikzlibrary{arrows.meta}
\usepackage[hidelinks,breaklinks]{hyperref}

\title{FA-BENCH: A BENCHMARK FOR PHONE- AND WORD-LEVEL TIMESTAMP ACCURACY\\
IN FORCED ALIGNMENT AND ASR ON CLEAN AND NOISY SPEECH}

\name{\begin{tabular}{@{}>{\fontsize{11pt}{13.2pt}\selectfont}c@{}}
Wei Chu$^{1}$, Yuanzhe Dong$^{1,2}$, Ke Tan$^{1}$, Dong Han$^{1}$, Yichao Zhou$^{3}$, Ruchao Fan$^{4}$, Bingshen Mu$^{5}$\\
Jingbei Li$^{7}$, Vishwas Shetty$^{6}$, Sarthak Bisht$^{8}$, Ziyue Qiu$^{8}$, Massa Baali$^{8}$, Rita Singh$^{8}$, Bhiksha Raj$^{8}$
\end{tabular}}
\address{\fontsize{11pt}{13.2pt}\selectfont%
  $^{1}$Olewave, USA\enspace
  $^{2}$Stanford University\enspace
  $^{3}$Cartesia\enspace
  $^{4}$Microsoft\enspace
  $^{5}$Northwestern Polytechnical University\\
  \fontsize{11pt}{13.2pt}\selectfont%
  $^{6}$University of California, Los Angeles\enspace
  $^{7}$Tsinghua University\enspace
  $^{8}$Carnegie Mellon University}

\newcommand{\recfn}{\textsuperscript{\ref{fn:records}}}

\begin{document}
\maketitle

\begin{abstract}
\looseness=-1
Forced alignment estimates the timestamps of each word, phone or character in
speech given its transcript. Published comparisons normalize transcripts, split the
data and match boundaries differently, so their numbers cannot be read
together. We present \textbf{FA-Bench}, an open framework that fixes those
choices once and releases the code, splits, phone mapping, text normalization
and scoring script, with results published periodically. Track~1 gives every
aligner the reference transcript and Track~2 gives it a recognizer's output,
on the same audio, clean and degraded four ways, with 21 open models
and 9 commercial APIs under a unified protocol. We score every boundary of an
utterance and check the two labels beside it, so a word the recognizer missed
or invented is charged. Using a tolerance-based $F_1$ as our primary metric
eliminates the 9\% to 14\% score inflation that standard MAE causes on
recognition-dependent systems in conversational speech. We then group boundaries by position and
how many adjacent words were recognized correctly, which shows where a system
lost the score. We discovered systematic bias in how current systems time
words, with Whisper about 150\,ms early and several commercial ASR APIs over
50\,ms late. Code and results are at
\url{https://github.com/olewave/fa-bench}.
\end{abstract}


\section{Introduction}
\label{sec:intro}

Given a transcript, forced alignment estimates where each of its words, phones
or characters starts and ends in the speech audio. HMM-GMM systems such as P2FA~\cite{yuan2008p2fa},
Prosodylab-Aligner~\cite{gorman2011prosodylab} and MFA~\cite{mcauliffe2017mfa}
are still in use, and neural frame
classifiers~\cite{zhu2022charsiu,kelley2024maps}, CTC aligners over
self-supervised encoders~\cite{baevski2020wav2vec2,rehman2025bfa}, attention
models~\cite{li2022neufa} and end-to-end differentiable
aligners~\cite{rousso2026falcon} have followed. Timestamped ASR
systems~\cite{bain2023whisperx,radford2023whisper,zusag2024crisperwhisper}
produce times for the words they decode, and have been compared only within
the Whisper family~\cite{zusag2024crisperwhisper,yeh2025aligner}.

Their published numbers cannot be read against each other, because every
evaluation normalizes transcripts, splits the data and matches boundaries its
own way, and most use clean speech. Some treat another aligner's
boundaries as ground truth~\cite{bain2023whisperx,zusag2024crisperwhisper,yeh2025aligner,rastorgueva2023nfa,ramirez2024universal1,qwen2026asr,mu2026llmfa}. Buckeye ships no
official split, so every paper picks its own~\cite{kelley2024maps,li2022neufa,rousso2026falcon}, and one
paper's held-out speaker is another's training speaker.

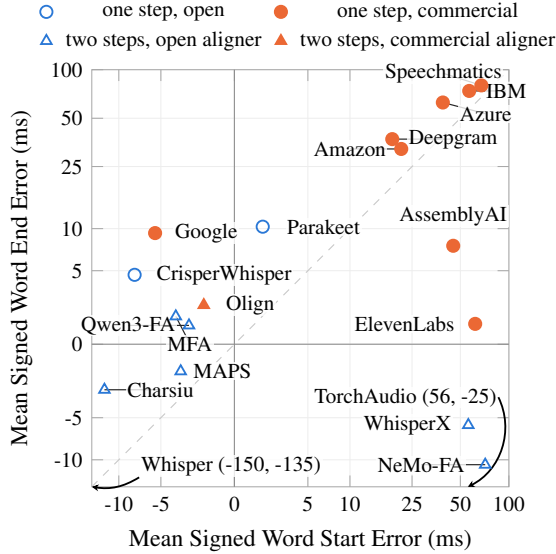
\begin{figure}[!t]
\centering
\definecolor{bias2a78d6}{HTML}{2a78d6}
\definecolor{biaseb6834}{HTML}{eb6834}
\definecolor{bias2a78d6}{HTML}{2a78d6}
\definecolor{biaseb6834}{HTML}{eb6834}
\begin{tikzpicture}
\begin{axis}[
  width=157pt, height=157pt,
  scale only axis, clip=false,
  xmin=-2.0322, xmax=3.9124, ymin=-2.0322, ymax=3.9124,
  xlabel={Mean Signed Word Start Error (ms)},
  ylabel={Mean Signed Word End Error (ms)},
  xlabel style={font=\small}, ylabel style={font=\small},
  xticklabel style={font=\footnotesize},
  yticklabel style={font=\footnotesize},
  xtick={-1.6472,-1.0476,0.0000,1.0476,1.6472,2.5321,3.2205,3.9124},
  xticklabels={-10,-5,0,5,10,25,50,100},
  ytick={-1.6472,-1.0476,0.0000,1.0476,1.6472,2.5321,3.2205,3.9124},
  yticklabels={-10,-5,0,5,10,25,50,100},
  axis line style={draw=black!35, line width=0.3pt},
  tick style={draw=black!35},
  grid=major, grid style={draw=black!7, line width=0.3pt},
  legend style={font=\footnotesize, at={(0.5,1.02)}, anchor=south,
                legend columns=2, draw=none, fill=none, column sep=3pt},
]
\addplot[draw=black!22, line width=0.4pt, dashed, forget plot] coordinates {(-2.0322,-2.0322) (3.9124,3.9124)};
\addplot[draw=black!45, line width=0.4pt, forget plot] coordinates {(-2.0322,0) (3.9124,0)};
\addplot[draw=black!45, line width=0.4pt, forget plot] coordinates {(0,-2.0322) (0,3.9124)};
\draw[draw=black!85, line width=0.45pt] (axis cs:-1.852,-0.649) -- (axis cs:-1.052,-0.649);
\node[font=\footnotesize, inner sep=0.5pt, fill=white, fill opacity=0.8, text opacity=1] at (axis cs:-1.052,-0.649) {Charsiu};
\node[font=\footnotesize, inner sep=0.5pt, fill=white, fill opacity=0.8, text opacity=1] at (axis cs:-0.145,0.989) {CrisperWhisper};
\node[font=\footnotesize, inner sep=0.5pt, fill=white, fill opacity=0.8, text opacity=1] at (axis cs:-0.399,1.584) {Google};
\draw[draw=black!85, line width=0.45pt] (axis cs:-0.835,0.397) -- (axis cs:-0.635,0.010);
\node[font=\footnotesize, inner sep=0.5pt, fill=white, fill opacity=0.8, text opacity=1] at (axis cs:-0.635,0.010) {MFA};
\node[font=\footnotesize, inner sep=0.5pt, fill=white, fill opacity=0.8, text opacity=1] at (axis cs:-0.170,-0.386) {MAPS};
\draw[draw=black!85, line width=0.45pt] (axis cs:-0.647,0.268) -- (axis cs:-1.515,0.268);
\node[font=\footnotesize, inner sep=0.5pt, fill=white, fill opacity=0.8, text opacity=1] at (axis cs:-1.515,0.268) {Qwen3-FA};
\node[font=\footnotesize, inner sep=0.5pt, fill=white, fill opacity=0.8, text opacity=1] at (axis cs:0.227,0.558) {Olign};
\node[font=\footnotesize, inner sep=0.5pt, fill=white, fill opacity=0.8, text opacity=1] at (axis cs:1.272,1.675) {Parakeet};
\draw[draw=black!85, line width=0.45pt] (axis cs:2.250,2.923) -- (axis cs:3.119,2.923);
\node[font=\footnotesize, inner sep=0.5pt, fill=white, fill opacity=0.8, text opacity=1] at (axis cs:3.119,2.923) {Deepgram};
\draw[draw=black!85, line width=0.45pt] (axis cs:2.378,2.783) -- (axis cs:1.647,2.783);
\node[font=\footnotesize, inner sep=0.5pt, fill=white, fill opacity=0.8, text opacity=1] at (axis cs:1.647,2.783) {Amazon};
\draw[draw=black!85, line width=0.45pt] (axis cs:2.971,3.445) -- (axis cs:3.581,3.285);
\node[font=\footnotesize, inner sep=0.5pt, fill=white, fill opacity=0.8, text opacity=1] at (axis cs:3.581,3.285) {Azure};
\node[font=\footnotesize, inner sep=0.5pt, fill=white, fill opacity=0.8, text opacity=1] at (axis cs:3.119,1.824) {AssemblyAI};
\node[font=\footnotesize, inner sep=0.5pt, fill=white, fill opacity=0.8, text opacity=1] at (axis cs:2.466,-1.152) {WhisperX};
\draw[draw=black!85, line width=0.45pt] (axis cs:3.347,3.610) -- (axis cs:3.874,3.610);
\node[font=\footnotesize, inner sep=0.5pt, fill=white, fill opacity=0.8, text opacity=1] at (axis cs:3.874,3.610) {IBM};
\node[font=\footnotesize, inner sep=0.5pt, fill=white, fill opacity=0.8, text opacity=1] at (axis cs:2.428,0.290) {ElevenLabs};
\draw[draw=black!85, line width=0.45pt] (axis cs:3.520,3.689) -- (axis cs:2.980,3.860);
\node[font=\footnotesize, inner sep=0.5pt, fill=white, fill opacity=0.8, text opacity=1] at (axis cs:2.980,3.860) {Speechmatics};
\draw[draw=black!85, line width=0.45pt] (axis cs:3.575,-1.717) -- (axis cs:2.662,-1.717);
\node[font=\footnotesize, inner sep=0.5pt, fill=white, fill opacity=0.8, text opacity=1] at (axis cs:2.662,-1.717) {NeMo-FA};
\node[font=\footnotesize, inner sep=0.5pt, fill=white, fill opacity=0.8, text opacity=1] (offlab0) at (axis cs:-0.048,-1.735) {Whisper (-150, -135)};
\draw[draw=black, line width=0.7pt, -{Stealth[length=2.6pt,width=2.6pt]}] (offlab0.south west) to[bend left=20] (axis cs:-2.0322,-2.0322);
\node[font=\footnotesize, inner sep=0.5pt, fill=white, fill opacity=0.8, text opacity=1] (offlab1) at (axis cs:2.440,-0.755) {TorchAudio (56, -25)};
\draw[draw=black, line width=0.7pt, -{Stealth[length=2.6pt,width=2.6pt]}] (offlab1.east) to[bend left=45] (axis cs:3.326,-2.0322);
\addplot[only marks, mark=o, mark size=2.3pt, draw=bias2a78d6, fill=white, line width=0.7pt] coordinates {(-1.422,0.989) (0.404,1.675)};
\addlegendentry{one step, open}
\addplot[only marks, mark=*, mark size=2.3pt, draw=biaseb6834, fill=biaseb6834, line width=0.7pt] coordinates {(-1.130,1.584) (2.250,2.923) (2.378,2.783) (2.971,3.445) (3.119,1.403) (3.347,3.610) (3.432,0.290) (3.520,3.689)};
\addlegendentry{one step, commercial}
\addplot[only marks, mark=triangle, mark size=2.3pt, draw=bias2a78d6, fill=white, line width=0.7pt] coordinates {(-1.852,-0.649) (-0.835,0.397) (-0.765,-0.386) (-0.647,0.268) (3.334,-1.152) (3.575,-1.717)};
\addlegendentry{two steps, open aligner}
\addplot[only marks, mark=triangle*, mark size=2.3pt, draw=biaseb6834, fill=biaseb6834, line width=0.7pt] coordinates {(-0.436,0.558)};
\addlegendentry{two steps, commercial aligner}
\end{axis}
\end{tikzpicture}
\caption{\small Mean signed word-boundary error on TIMIT test, hypothesis minus reference, over the words the label alignment matched. Track~2 only. Every two-step point aligns Qwen3-ASR's transcript, so the recognizer is fixed and only the aligner varies. Both axes are warped by $\mathrm{asinh}(v/4)$. The distance from the dashed line $y=x$ is the mean error in word duration.}
\label{fig:bias-timit-core_test-word}
\vspace{5pt}
\end{figure}

We build FA-Bench for word and phone forced alignment, and also for evaluating the timestamps output from ASR
systems and forced alignment using ASR results as reference words. In practice the
words to align usually come from a recognizer. We make four
contributions.
\begin{itemize}[label=$\bullet$,leftmargin=0.8em,labelsep=0.3em,itemsep=1pt,topsep=1pt,parsep=0pt]
\item \textbf{An open framework for researchers to exactly reproduce results
and for developers to objectively compare systems.} For researchers, we
release the code, splits, phone mapping, text normalization and scoring
script, publish benchmark results periodically, and a new system is one
wrapper away. For developers choosing a service or model from many that come with no
evaluation, the published \texttt{records/}\recfn{} are where to look, and
the harness runs on their own proprietary data.

\item \textbf{Scoring every boundary and examining the two labels adjacent to
it.} MFA's evaluation~\cite{mcauliffe2026eval} assumes a gold transcript and never
checks the labels beside a boundary, so once the words come from ASR, a
boundary between two misread units still counts. It also skips the first and
last boundary of each utterance, 18--22\% of the word boundaries here. Rousso
et al.~\cite{rousso2024tradition} and AssemblyAI~\cite{ramirez2024universal1}
scored matched words only, so a recognizer pays nothing for a word it drops.
Yeh et al.~\cite{yeh2025aligner} score every word but check only its end,
which misses a late start.

\item \textbf{Grouping boundaries into five categories for analyzing where an
ASR's timestamps lost the score.} A single $F_1$ or MAE cannot tell a system that misreads words from one that
times correct words badly. It also averages utterance edges with
interior boundaries as if they were one problem. An edge is found against
silence, and an interior boundary sits between two words. So we class each boundary by where it
sits and how many adjacent words match the reference.

\item \textbf{Discovering significant systematic bias in the word start and
end times of current open systems and APIs.}
Fig.~\ref{fig:bias-timit-core_test-word} shows it and Sec.~\ref{sec:results}
discusses it.

\end{itemize}

\section{The benchmark}

\label{sec:bench}

Every cell of every system is published in
\texttt{records/}\footnote{\label{fn:records}Full records are at \url{https://github.com/olewave/fa-bench/tree/main/records}}
with its metrics.

\subsection{Tasks}
\label{sec:tasks}

FA-Bench evaluates a word tier and a phone tier under two tracks. Track~1
measures timing alone and Track~2 timing plus whatever the recognizer got
wrong, so the two blocks answer different questions.

\begin{center}
\vspace{-4pt}
\small
\setlength{\tabcolsep}{4pt}
\begin{tabular}{@{}lccc@{}}
\toprule
& \textbf{Track~1} & \multicolumn{2}{c}{\textbf{Track~2}}\\
\cmidrule(lr){2-2}\cmidrule(lr){3-4}
& reference  & one step              & two steps \\
& transcript & ASR with timestamps & align after ASR \\
\midrule
Word tier   & \checkmark & \checkmark & \checkmark \\
Phone tier  & \checkmark & $\times$   & \checkmark\recfn \\
\bottomrule
\end{tabular}
\vspace{-4pt}
\end{center}

\noindent\textbf{Open and closed systems.}
The versions and settings of the 21 open models and the 9 APIs are
documented in \texttt{records/}\recfn. The company of the first group of authors offers
Olign API~\cite{olewave2024olign}, which is not trained with dev or test
splits of TIMIT and Buckeye.

\subsection{Data}
\label{sec:data}

\textbf{Corpora.}
We use TIMIT~\cite{garofolo1993timit} for read speech and
Buckeye~\cite{pitt2007buckeye} for conversational speech, both with word and
phone boundaries marked by hand. Phone tiers were annotated from the
audio rather than expanded from pronunciation dictionaries. In Buckeye, the
canonical and realized pronunciations differ for 58.7\% of word tokens.
A dictionary-based aligner therefore proposes sounds the talker never made.

\textbf{Splits.}
All evaluation speakers are held out. TIMIT uses the corpus core test set and
Kaldi's development split. Buckeye has no official split, so we define
one while preserving its sex$\times$age structure. Each of the four
younger/older $\times$ female/male cells contributes six speakers to training
and two each to development and test.

\textbf{Transcript normalization.}
TIMIT \texttt{.WRD} transcripts are lower-cased without further modification.
Buckeye sometimes hides a spoken word inside an event tag such as
\texttt{<LAUGH-here>}. We take the word back when the phone tier under
the tag holds real speech, keeping the annotated times, and drop the tag
when it does not. Because aligners may renormalize their
input to match their lexicons, their outputs are mapped back to the input
tokens whenever the letters spell that token back, ignoring case and
punctuation.

\textbf{Degradations.}
We follow the Kaldi~\cite{povey2011kaldi} VoxCeleb recipe and its default
parameters, which add MUSAN~\cite{snyder2015musan} noise, music and babble, all of them
recorded, together with simulated room impulse responses~\cite{ko2017rirs}.

\subsection{Scoring}
\label{sec:protocol}

\textbf{Shared phone inventory.}
Each system aligns in its own phone set, and the two corpora do not share one
either. We map its phones and the reference phones into
TIMIT-39~\cite{lee1989timit39} afterwards, the conventional reduction for
these corpora and the one Kaldi ships~\cite{povey2011kaldi}. The map exists
only so the rows can be read against each other. It renames a unit and never
moves a boundary, and the only label we drop is the glottal stop, 1.6\% of
TIMIT phones and none of Buckeye's. The source alphabets and the mappings are
in \texttt{records/}\recfn. What costs these systems is the segmentation
convention, e.g. \texttt{bcl} before \texttt{b}.
Although we reported numbers with the closure mapped as silence, FA-Bench
as a tool still offers users the option of merging the closure into its
burst.

\textbf{Metrics.}
In FA-Bench, a \emph{matched boundary} is defined as a detected boundary whose
adjacent labels match the ground truth labels. The label can be word or phone.
Note that for Track~1, where the reference words are available, every word
boundary is a matched boundary. For Track~2, the matched boundaries are obtained by
aligning recognized and reference words first. Then, the \emph{boundary mean
absolute error (MAE)} is calculated by averaging the absolute difference in ms
between their times and the ground truth. In Track~2, due to recognition
errors, not all ground truth boundaries are used in the MAE calculation.
Qwen3-ASR makes 13.5\% word errors on noisy Buckeye, so its MAE is computed on
83\% of the reference boundaries.

A recognizer can drop the words it finds hard and gain from doing so, since
MAE then reports only the easy words it kept. Scored only at the words a
Track~2 recognizer got right, Olign's Buckeye MAE drops by 9\% to 14\%, while
its TIMIT MAE does not move. Pitch tracking has the same trade-off, where a
low gross pitch error can come with a high voicing error, so F0 frame error
counts both~\cite{chu2009ffe}. To count every reference
boundary we use \emph{boundary $F_1$}, whose recall takes every reference
boundary and whose precision takes every hypothesized one. A boundary is a hit
only when it is a matched boundary and its time falls within the tolerance of
the ground truth. The two utterance edges are boundaries as well, against
silence, and silence matches, so an edge answers to its one word. A word the
system misread therefore costs it the two boundaries beside that word.
An utterance the system returns nothing for costs it every boundary in that
utterance, and every word and phone in it counts as deleted.
Tables~\ref{tab:main} and~\ref{tab:phone} report 20\,ms. The tolerance is
swept from 10 to 100\,ms in \texttt{records/}\recfn, with over-segmentation
(OS) and R-value~\cite{rasanen2009rvalue}.

\textbf{Grouped boundaries.} We wonder how recognition errors affect the
accuracy of forced alignment, which includes what a single mismatched label
costs against two matched labels, whether an utterance edge is harder than an
interior boundary, whether finding where speech begins differs from finding
where it ends, and what is left once both labels are wrong. So we group every
boundary by where it sits and whether its adjacent labels match, into 5
categories (Table~\ref{tab:class}).

\section{Results}
\newlength{\fabdgt}

\begin{table}[!t]
\centering
\renewcommand{\arraystretch}{1.0}
\setlength{\arrayrulewidth}{0.25pt}
\setlength{\aboverulesep}{0pt}\setlength{\belowrulesep}{0pt}
\caption{\small Word-tier boundary error on the test split of each corpus, on clean audio and under degradation, \emph{noisy} being the mean of the four. Every number here comes from our own runs. MAE is over word boundaries in ms. $F_1$ is over every word boundary, the two utterance edges included. A boundary is a hit only when the words on each side of it match the reference and the time falls within the tolerance of the reference boundary. The superscript on $F_1$ is that tolerance in ms. The step size of a word boundary of each system is shown in Table~\ref{tab:class}. Open means the weights are published, and NeMo-FA, WhisperX, TorchAudio, BFA, MMS-FA, UnitY2, Charsiu, MFA and MAPS also publish the code that trained them. The NeuFA model is trained with its open-source code on GitHub and is not released. W is Whisper large-v3, P and Parakeet are Parakeet-TDT, Q and Qwen3 are Qwen3-ASR, and G is Google Chirp~2. Whisper-ts is Whisper-timestamped and NeMo-FA uses its Conformer checkpoint. The APIs are Speechmatics Enhanced, Deepgram Nova-3, IBM Large US English, Azure en-US, AssemblyAI Universal~3.5 Pro, Amazon Transcribe en-US, ElevenLabs Scribe~v2, Google Chirp~2 and Olign~0.9. MFA is Montreal Forced Aligner 3.4. MAPS and NeuFA trained on 7 of 8 speakers in both Buckeye splits.}
\label{tab:main}
\fontsize{8pt}{8.8pt}\selectfont
\settowidth{\fabdgt}{0}
\setlength{\tabcolsep}{0.05pt}
\begin{tabular*}{\columnwidth}{@{}l@{\hspace{0.2pt}}>{\fontsize{8pt}{8.8pt}\selectfont}l|@{\extracolsep{\fill}} >{\hspace{\dimexpr-0pt\relax}}c<{\hspace{\dimexpr0.3pt+0pt\relax}}|>{\hspace{\dimexpr0.3pt\relax}}c<{\hspace{\dimexpr0.2pt\relax}}|>{\hspace{\dimexpr0.2pt\relax}}c<{\hspace{\dimexpr0.7pt\relax}}|>{\hspace{\dimexpr0.7pt-0pt\relax}}c<{\hspace{\dimexpr0.3pt+0pt\relax}}|>{\hspace{\dimexpr0.3pt\relax}}c<{\hspace{\dimexpr0.2pt\relax}}|>{\hspace{\dimexpr0.2pt\relax}}c<{\hspace{\dimexpr0.7pt\relax}}|>{\hspace{\dimexpr0.7pt-0pt\relax}}c<{\hspace{\dimexpr0.3pt+0pt\relax}}|>{\hspace{\dimexpr0.3pt\relax}}c<{\hspace{\dimexpr0.2pt\relax}}|>{\hspace{\dimexpr0.2pt\relax}}c<{\hspace{\dimexpr0.7pt\relax}}|>{\hspace{\dimexpr0.7pt-0pt\relax}}c<{\hspace{\dimexpr0.3pt+0pt\relax}}|>{\hspace{\dimexpr0.3pt\relax}}c<{\hspace{\dimexpr0.2pt\relax}}|>{\hspace{\dimexpr0.2pt\relax}}c@{}}
\toprule
\multirow{3}{*}{\rotatebox[origin=c]{90}{\fontsize{5.5pt}{6pt}\selectfont Family}} & & \multicolumn{6}{c|}{TIMIT test} & \multicolumn{6}{c}{Buckeye test}\\
\cmidrule(lr){3-8}\cmidrule(lr){9-14}
 & System & \multicolumn{3}{c|}{clean} & \multicolumn{3}{c|}{noisy} & \multicolumn{3}{c|}{clean} & \multicolumn{3}{c}{noisy}\\
\cmidrule(lr){3-5}\cmidrule(lr){6-8}\cmidrule(lr){9-11}\cmidrule(lr){12-14}
& \multicolumn{13}{l}{\small\textbf{Track~1}\emph{, given the reference transcript}}\\
& & \scalebox{0.9}[1]{MAE} & \scalebox{0.74}[1]{\fontsize{6pt}{6.6pt}\selectfont$\boldsymbol{F_1^{\scriptscriptstyle 20}}$} & \scalebox{0.74}[1]{\fontsize{6pt}{6.6pt}\selectfont$\boldsymbol{F_1^{\scriptscriptstyle 50}}$} & \scalebox{0.9}[1]{MAE} & \scalebox{0.74}[1]{\fontsize{6pt}{6.6pt}\selectfont$\boldsymbol{F_1^{\scriptscriptstyle 20}}$} & \scalebox{0.74}[1]{\fontsize{6pt}{6.6pt}\selectfont$\boldsymbol{F_1^{\scriptscriptstyle 50}}$} & \scalebox{0.9}[1]{MAE} & \scalebox{0.74}[1]{\fontsize{6pt}{6.6pt}\selectfont$\boldsymbol{F_1^{\scriptscriptstyle 20}}$} & \scalebox{0.74}[1]{\fontsize{6pt}{6.6pt}\selectfont$\boldsymbol{F_1^{\scriptscriptstyle 50}}$} & \scalebox{0.9}[1]{MAE} & \scalebox{0.74}[1]{\fontsize{6pt}{6.6pt}\selectfont$\boldsymbol{F_1^{\scriptscriptstyle 20}}$} & \scalebox{0.74}[1]{\fontsize{6pt}{6.6pt}\selectfont$\boldsymbol{F_1^{\scriptscriptstyle 50}}$}\\
\midrule
\multirow{13}{*}{\smash{\rotatebox[origin=c]{90}{Open}}} & NeMo-FA\,{\tiny\cite{rastorgueva2023nfa}} & \phantom{0}47.4 & .17 & \phantom{0}.41 & \phantom{0}47.1 & .18 & \phantom{0}.43 & \phantom{0}62.8 & .21 & \phantom{0}.49 & \phantom{0}65.1 & .21 & \phantom{0}.50\\
                                   & stable-ts\,{\tiny\cite{jianfch2023stablets}} & \phantom{0}96.8 & .18 & \phantom{0}.39 & \phantom{0}94.5 & .19 & \phantom{0}.41 & \phantom{0}71.2 & .29 & \phantom{0}.55 & \phantom{0}75.2 & .30 & \phantom{0}.55\\
                                   & WhisperX\,{\tiny\cite{bain2023whisperx}} & \phantom{0}37.2 & .18 & \phantom{0}.43 & \phantom{0}39.5 & .17 & \phantom{0}.42 & \phantom{0}41.7 & .16 & \phantom{0}.49 & \phantom{0}52.0 & .17 & \phantom{0}.48\\
                                   & TorchAudio\,{\tiny\cite{yang2022torchaudio}} & \phantom{0}39.8 & .21 & \phantom{0}.47 & \phantom{0}42.4 & .19 & \phantom{0}.45 & \phantom{0}42.4 & .18 & \phantom{0}.53 & \phantom{0}51.4 & .18 & \phantom{0}.51\\
                                   & BFA\,{\tiny\cite{rehman2025bfa}} & \phantom{0}51.0 & .27 & \phantom{0}.62 & \phantom{0}56.7 & .24 & \phantom{0}.57 & \phantom{0}57.2 & .30 & \phantom{0}.60 & \phantom{0}61.7 & .29 & \phantom{0}.58\\
                                   & MMS-FA\,{\tiny\cite{pratap2024mms}} & \phantom{0}31.1 & .28 & \phantom{0}.64 & \phantom{0}31.5 & .26 & \phantom{0}.62 & \phantom{0}37.1 & .23 & \phantom{0}.64 & \phantom{0}41.2 & .22 & \phantom{0}.63\\
                                   & UnitY2\,{\tiny\cite{seamless2023}} & \phantom{0}52.2 & .34 & \phantom{0}.69 & \phantom{0}53.8 & .35 & \phantom{0}.68 & \phantom{0}40.6 & .48 & \phantom{0}.79 & \phantom{0}49.0 & .47 & \phantom{0}.78\\
                                   & Qwen3-FA\,{\tiny\cite{qwen2026asr,mu2026llmfa}} & \phantom{0}32.2 & .40 & \phantom{0}.82 & \phantom{0}38.5 & .38 & \phantom{0}.78 & \phantom{0}33.8 & .43 & \phantom{0}.84 & \phantom{0}50.8 & .39 & \phantom{0}.77\\
                                   & CrisperW\,{\tiny\cite{zusag2024crisperwhisper}} & \phantom{0}34.1 & .44 & \phantom{0}.79 & \phantom{0}36.9 & .43 & \phantom{0}.77 & \phantom{0}43.1 & .49 & \phantom{0}.78 & \phantom{0}52.5 & .44 & \phantom{0}.74\\
                                   & NeuFA\,{\tiny\cite{li2022neufa}} & \phantom{0}56.9 & .52 & \phantom{0}.70 & \phantom{0}88.6 & .45 & \phantom{0}.64 & \phantom{0}33.9 & .70 & \phantom{0}.86 & \phantom{0}59.1 & .60 & \phantom{0}.79\\
                                   & Charsiu\,{\tiny\cite{zhu2022charsiu}} & \phantom{0}28.1 & .54 & \phantom{0}.86 & \phantom{0}40.2 & .49 & \phantom{0}.79 & \phantom{0}28.1 & .61 & \phantom{0}.87 & \phantom{0}60.3 & .49 & \phantom{0}.74\\
                                   & MFA\,{\tiny\cite{mcauliffe2017mfa}} & \phantom{0}21.8 & .64 & \phantom{0}\textbf{.92} & \phantom{0}\textbf{30.2} & .62 & \phantom{0}\textbf{.86} & \phantom{0}\textbf{21.3} & .68 & \phantom{0}\textbf{.93} & \phantom{0}\textbf{35.6} & .58 & \phantom{0}.83\\
                                   & MAPS\,{\tiny\cite{kelley2024maps}} & \phantom{0}25.5 & .71 & \phantom{0}.89 & 110.0 & .40 & \phantom{0}.59 & \phantom{0}37.9 & .62 & \phantom{0}.84 & 111.4 & .44 & \phantom{0}.68\\
\cmidrule(lr){1-14}
{\tiny API}                        & Olign\,{\tiny\cite{olewave2024olign}} & \phantom{0}\textbf{18.2} & \textbf{.78} & \phantom{0}.91 & \phantom{0}33.0 & \textbf{.67} & \phantom{0}.82 & \phantom{0}22.5 & \textbf{.74} & \phantom{0}.91 & \phantom{0}36.5 & \textbf{.66} & \phantom{0}\textbf{.85}\\
\midrule
& \multicolumn{13}{l}{\small\textbf{Track~2}\emph{, one step. The ASR times its own words}}\\
& & \scalebox{0.9}[1]{MAE} & \scalebox{0.74}[1]{\fontsize{6pt}{6.6pt}\selectfont$\boldsymbol{F_1^{\scriptscriptstyle 20}}$} & \scalebox{0.78}[1]{WER} & \scalebox{0.9}[1]{MAE} & \scalebox{0.74}[1]{\fontsize{6pt}{6.6pt}\selectfont$\boldsymbol{F_1^{\scriptscriptstyle 20}}$} & \scalebox{0.78}[1]{WER} & \scalebox{0.9}[1]{MAE} & \scalebox{0.74}[1]{\fontsize{6pt}{6.6pt}\selectfont$\boldsymbol{F_1^{\scriptscriptstyle 20}}$} & \scalebox{0.78}[1]{WER} & \scalebox{0.9}[1]{MAE} & \scalebox{0.74}[1]{\fontsize{6pt}{6.6pt}\selectfont$\boldsymbol{F_1^{\scriptscriptstyle 20}}$} & \scalebox{0.78}[1]{WER}\\
\midrule
\multirow{5}{*}{\smash{\rotatebox[origin=c]{90}{Open}}} & Whisper\,{\tiny\cite{radford2023whisper}} & 148.0 & .10 & \phantom{0}2.9 & 151.8 & .10 & \phantom{0}3.2 & 122.7 & .14 & 13.7 & 123.0 & .13 & 17.5\\
                                   & Whisper-ts\,{\tiny\cite{louradour2023whisperts}} & 161.6 & .11 & \phantom{0}2.5 & 163.9 & .11 & \phantom{0}3.4 & 133.3 & .17 & 13.4 & 136.0 & .16 & 17.9\\
                                   & TorchAudio               & \phantom{0}37.8 & .17 & 10.8 & \phantom{0}38.2 & .12 & 25.7 & \phantom{0}37.8 & .08 & 27.8 & \phantom{0}39.2 & .05 & 47.0\\
                                   & Parakeet\,{\tiny\cite{xu2023tdt}} & \phantom{0}79.3 & .18 & \phantom{0}2.4 & \phantom{0}77.1 & .18 & \phantom{0}3.5 & \phantom{0}80.7 & .19 & 11.0 & \phantom{0}80.5 & .18 & 15.3\\
                                   & CrisperW                 & \phantom{0}33.5 & .42 & \phantom{0}3.6 & \phantom{0}36.2 & .41 & \phantom{0}4.3 & \phantom{0}37.6 & .43 & 11.5 & \phantom{0}41.0 & .37 & 15.5\\
\cmidrule(lr){1-14}
\multirow{8}{*}{\smash{\rotatebox[origin=c]{90}{API}}} & Azure\,{\tiny\cite{microsoft2026azure}} & \phantom{0}64.4 & .14 & \phantom{0}3.4 & \phantom{0}67.4 & .14 & \phantom{0}4.6 & \phantom{0}84.7 & .14 & 11.5 & \phantom{0}92.0 & .13 & 15.8\\
                                   & IBM\,{\tiny\cite{ibm2026watson}} & \phantom{0}72.4 & .15 & \phantom{0}5.1 & \phantom{0}75.9 & .14 & \phantom{0}9.2 & \phantom{0}62.0 & .18 & 12.4 & \phantom{0}65.0 & .16 & 19.9\\
                                   & Speechmatics\,{\tiny\cite{speechmatics2026batch}} & \phantom{0}77.1 & .15 & \phantom{0}2.5 & \phantom{0}81.5 & .14 & \phantom{0}3.4 & \phantom{0}74.2 & .14 & 11.4 & \phantom{0}81.6 & .13 & 14.4\\
                                   & ElevenLabs\,{\tiny\cite{elevenlabs2026scribe}} & \phantom{0}32.1 & .17 & \phantom{0}2.0 & \phantom{0}32.6 & .16 & \phantom{0}3.4 & \phantom{0}34.8 & .11 & 10.6 & \phantom{0}35.9 & .11 & 14.1\\
                                   & Deepgram\,{\tiny\cite{deepgram2026nova3}} & \phantom{0}67.6 & .21 & \phantom{0}2.4 & \phantom{0}70.2 & .20 & \phantom{0}4.4 & \phantom{0}97.8 & .20 & 10.8 & \phantom{0}93.3 & .18 & 16.8\\
                                   & AssemblyAI\,{\tiny\cite{assemblyai2026universal}} & \phantom{0}47.9 & .24 & \phantom{0}3.7 & \phantom{0}53.9 & .22 & \phantom{0}4.2 & \phantom{0}56.4 & .20 & 12.2 & \phantom{0}66.4 & .18 & 15.3\\
                                   & Amazon\,{\tiny\cite{amazon2026transcribe}} & \phantom{0}55.7 & .25 & \phantom{0}2.7 & \phantom{0}56.6 & .26 & \phantom{0}2.9 & \phantom{0}52.8 & .29 & 10.3 & \phantom{0}53.7 & .27 & 12.3\\
                                   & Google\,{\tiny\cite{google2026chirp2}} & \phantom{0}32.7 & .44 & \phantom{0}4.2 & \phantom{0}32.6 & .45 & \phantom{0}6.7 & \phantom{0}26.9 & .44 & 13.1 & \phantom{0}\textbf{28.1} & .41 & 18.5\\
& \multicolumn{13}{l}{\quad\small\emph{two steps. $\boldsymbol{\to}$ denotes ASR then aligner}}\\
\multirow{3}{*}{\smash{\rotatebox[origin=c]{90}{Open}}} & W$\to$WhisperX           & \phantom{0}35.1 & .18 & \phantom{0}2.9 & \phantom{0}38.4 & .16 & \phantom{0}6.6 & \phantom{0}46.2 & .13 & 14.2 & \phantom{0}55.6 & .12 & 19.3\\
                                   & Q$\to$Qwen3-FA           & \phantom{0}32.6 & .39 & \phantom{0}2.0 & \phantom{0}38.3 & .37 & \phantom{0}2.4 & \phantom{0}30.6 & .39 & 10.2 & \phantom{0}41.7 & .34 & 13.5\\
                                   & Q$\to$MFA                & \phantom{0}21.8 & .63 & \phantom{0}2.0 & \phantom{0}\textbf{29.6} & \textbf{.61} & \phantom{0}2.8 & \phantom{0}22.7 & .60 & 10.4 & \phantom{0}33.9 & .50 & 16.5\\
\cmidrule(lr){1-14}
\multirow{3}{*}{\smash{\rotatebox[origin=c]{90}{API}}} & Q$\to$Olign              & \phantom{0}\textbf{18.1} & \textbf{.76} & \phantom{0}2.0 & \phantom{0}34.2 & .61 & \phantom{0}2.4 & \phantom{0}20.6 & .67 & 10.2 & \phantom{0}30.3 & \textbf{.55} & 13.6\\
                                   & G$\to$Olign              & \phantom{0}18.1 & .73 & \phantom{0}4.2 & \phantom{0}33.7 & .57 & \phantom{0}6.7 & \phantom{0}20.2 & .63 & 13.1 & \phantom{0}29.4 & .51 & 18.6\\
                                   & P$\to$Olign              & \phantom{0}18.2 & .75 & \phantom{0}2.4 & \phantom{0}33.8 & .60 & \phantom{0}3.5 & \phantom{0}\textbf{20.1} & \textbf{.67} & 10.7 & \phantom{0}29.5 & .55 & 15.2\\
\bottomrule
\end{tabular*}
\end{table}

\begin{table}[!t]
\centering
\renewcommand{\arraystretch}{1.0}
\setlength{\arrayrulewidth}{0.25pt}
\setlength{\aboverulesep}{0pt}\setlength{\belowrulesep}{0pt}
\caption{\small Phone-tier boundary error on the test split of each corpus, on clean audio and under degradation, \emph{noisy} being the mean of the four, laid out as Table~\ref{tab:main}. Every Track~1 aligner is given the reference words and derives its own phones, and PER is the phone error rate of what it derived against the gold phones, per split and condition. FALCON, daggered, is handed the gold phones, and overlaps our Buckeye split in its training data. We use the FALCON released model and did not retrain it. No one-step ASR in Table~\ref{tab:main} emits phone timestamps, so Track~2 has only the two-step rows whose aligner does, and their PER carries the recognizer's word errors too.}
\label{tab:phone}
\fontsize{8pt}{8.8pt}\selectfont
\settowidth{\fabdgt}{0}
\setlength{\tabcolsep}{0.05pt}
\begin{tabular*}{\columnwidth}{@{}l@{\hspace{0.2pt}}>{\fontsize{8pt}{8.8pt}\selectfont}l|@{\extracolsep{\fill}} >{\hspace{\dimexpr-0pt\relax}}c<{\hspace{\dimexpr0.3pt+0pt\relax}}|>{\hspace{\dimexpr0.3pt\relax}}c<{\hspace{\dimexpr0.2pt\relax}}|>{\hspace{\dimexpr0.2pt\relax}}c<{\hspace{\dimexpr0.7pt\relax}}|>{\hspace{\dimexpr0.7pt-0pt\relax}}c<{\hspace{\dimexpr0.3pt+0pt\relax}}|>{\hspace{\dimexpr0.3pt\relax}}c<{\hspace{\dimexpr0.2pt\relax}}|>{\hspace{\dimexpr0.2pt\relax}}c<{\hspace{\dimexpr0.7pt\relax}}|>{\hspace{\dimexpr0.7pt+0.5\fabdgt-0pt\relax}}c<{\hspace{\dimexpr0.3pt+0.5\fabdgt+0pt\relax}}|>{\hspace{\dimexpr0.3pt\relax}}c<{\hspace{\dimexpr0.2pt\relax}}|>{\hspace{\dimexpr0.2pt\relax}}c<{\hspace{\dimexpr0.7pt\relax}}|>{\hspace{\dimexpr0.7pt-0pt\relax}}c<{\hspace{\dimexpr0.3pt+0pt\relax}}|>{\hspace{\dimexpr0.3pt\relax}}c<{\hspace{\dimexpr0.2pt\relax}}|>{\hspace{\dimexpr0.2pt\relax}}c@{}}
\toprule
\multirow{3}{*}{\rotatebox[origin=c]{90}{\fontsize{5.5pt}{6pt}\selectfont Family}} & & \multicolumn{6}{c|}{TIMIT test} & \multicolumn{6}{c}{Buckeye test}\\
\cmidrule(lr){3-8}\cmidrule(lr){9-14}
 & System & \multicolumn{3}{c|}{clean} & \multicolumn{3}{c|}{noisy} & \multicolumn{3}{c|}{clean} & \multicolumn{3}{c}{noisy}\\
\cmidrule(lr){3-5}\cmidrule(lr){6-8}\cmidrule(lr){9-11}\cmidrule(lr){12-14}
& \multicolumn{13}{l}{\small\textbf{Track~1}\emph{, reference words, except FALCON}\smash{$^\dagger$}}\\
& & \scalebox{0.9}[1]{MAE} & \scalebox{0.74}[1]{\fontsize{6pt}{6.6pt}\selectfont$\boldsymbol{F_1^{\scriptscriptstyle 20}}$} & \scalebox{0.80}[1]{PER} & \scalebox{0.9}[1]{MAE} & \scalebox{0.74}[1]{\fontsize{6pt}{6.6pt}\selectfont$\boldsymbol{F_1^{\scriptscriptstyle 20}}$} & \scalebox{0.80}[1]{PER} & \scalebox{0.9}[1]{MAE} & \scalebox{0.74}[1]{\fontsize{6pt}{6.6pt}\selectfont$\boldsymbol{F_1^{\scriptscriptstyle 20}}$} & \scalebox{0.80}[1]{PER} & \scalebox{0.9}[1]{MAE} & \scalebox{0.74}[1]{\fontsize{6pt}{6.6pt}\selectfont$\boldsymbol{F_1^{\scriptscriptstyle 20}}$} & \scalebox{0.80}[1]{PER}\\
\midrule
\multirow{7}{*}{\smash{\rotatebox[origin=c]{90}{Open}}} & BFA\,{\tiny\cite{rehman2025bfa}} & 45.1 & .18 & 36.2 & 48.9 & .18 & 36.2 & 50.2 & .25 & 32.1 & \phantom{0}54.8 & .25 & 32.2\\
                                   & TorchAudio\,{\tiny\cite{yang2022torchaudio}} & 33.8 & .21 & 35.9 & 33.3 & .21 & 35.9 & 34.5 & .27 & 28.5 & \phantom{0}41.3 & .26 & 28.5\\
                                   & NeuFA\,{\tiny\cite{li2022neufa}} & 20.5 & .33 & 36.0 & 56.1 & .28 & 39.1 & 22.6 & .46 & 28.0 & \phantom{0}46.5 & .40 & 30.7\\
                                   & Charsiu\,{\tiny\cite{zhu2022charsiu}} & 19.9 & .36 & 30.3 & 28.3 & .35 & 31.1 & 21.1 & .42 & 31.7 & \phantom{0}50.4 & .36 & 34.8\\
                                   & MFA\,{\tiny\cite{mcauliffe2017mfa}} & 11.6 & .39 & 34.8 & 17.0 & .37 & 35.0 & 15.3 & .52 & 26.1 & \phantom{0}26.1 & .46 & 29.5\\
                                   & MAPS\,{\tiny\cite{kelley2024maps}} & 15.3 & .45 & 30.7 & 89.6 & .32 & 30.7 & 22.9 & .44 & 34.9 & 104.9 & .31 & 34.9\\
                                   & FALCON\smash{$^\dagger$}\,{\tiny\cite{rousso2026falcon}} & 22.3 & \textbf{.67} & \phantom{0}0.0 & 67.8 & \textbf{.52} & \phantom{0}0.0 & 40.6 & \textbf{.58} & \phantom{0}0.7 & 99.5 & .46 & \phantom{0}0.7\\
\cmidrule(lr){1-14}
{\tiny API}                        & Olign\,{\tiny\cite{olewave2024olign}} & \phantom{0}\textbf{8.4} & .41 & 35.4 & \textbf{13.1} & .39 & 35.5 & \textbf{12.7} & .52 & 28.2 & \phantom{0}\textbf{24.2} & \textbf{.48} & 28.5\\
\midrule
& \multicolumn{13}{l}{\small\textbf{Track~2}\emph{, two steps. $\boldsymbol{\to}$ denotes ASR then aligner}}\\
& & \scalebox{0.9}[1]{MAE} & \scalebox{0.74}[1]{\fontsize{6pt}{6.6pt}\selectfont$\boldsymbol{F_1^{\scriptscriptstyle 20}}$} & \scalebox{0.80}[1]{PER} & \scalebox{0.9}[1]{MAE} & \scalebox{0.74}[1]{\fontsize{6pt}{6.6pt}\selectfont$\boldsymbol{F_1^{\scriptscriptstyle 20}}$} & \scalebox{0.80}[1]{PER} & \scalebox{0.9}[1]{MAE} & \scalebox{0.74}[1]{\fontsize{6pt}{6.6pt}\selectfont$\boldsymbol{F_1^{\scriptscriptstyle 20}}$} & \scalebox{0.80}[1]{PER} & \scalebox{0.9}[1]{MAE} & \scalebox{0.74}[1]{\fontsize{6pt}{6.6pt}\selectfont$\boldsymbol{F_1^{\scriptscriptstyle 20}}$} & \scalebox{0.80}[1]{PER}\\
\midrule
{\tiny Open}                       & Q$\to$MFA                & 11.6 & .39 & 34.8 & 16.6 & .37 & 35.2 & 17.6 & .50 & 28.2 & \phantom{0}27.2 & .45 & 32.4\\
\cmidrule(lr){1-14}
\multirow{3}{*}{\smash{\rotatebox[origin=c]{90}{API}}} & Q$\to$Olign              & \phantom{0}7.5 & .41 & 35.4 & 12.8 & \textbf{.39} & 35.6 & 12.4 & .52 & 29.2 & \phantom{0}22.6 & \textbf{.46} & 31.0\\
                                   & G$\to$Olign              & \phantom{0}8.3 & .41 & 36.2 & 13.2 & .38 & 36.7 & 12.4 & .51 & 30.3 & \phantom{0}22.0 & .44 & 33.4\\
                                   & P$\to$Olign              & \phantom{0}\textbf{7.5} & \textbf{.41} & 35.4 & \textbf{12.7} & .39 & 35.9 & \textbf{12.3} & \textbf{.52} & 29.1 & \phantom{0}\textbf{21.8} & .46 & 31.8\\
\bottomrule
\end{tabular*}
\end{table}

\begin{table}[t]
\centering
\fontsize{8pt}{8.8pt}\selectfont
\setlength{\tabcolsep}{0.05pt}\renewcommand{\arraystretch}{1.0}
\setlength{\aboverulesep}{0pt}\setlength{\belowrulesep}{0pt}
\caption{\small The $F_1$s at 20\,ms of different types of word-tier boundaries of Track~2 on Buckeye test, on clean audio and under degradation, \emph{noisy} being the mean of the four. A boundary inside ($I$) an utterance has two adjacent words. We use $\mathrm{M2}_{I}$ to denote both words are matched and $\mathrm{M1}_{I}$ to denote one label is matched, where M denotes matched. $\mathrm{M1}_{B}$ and $\mathrm{M1}_{E}$ denote the beginning ($B$) and the ending ($E$) boundary of an utterance, each with only one adjacent word. $\mathrm{M0}$ denotes the remaining boundaries. The first row is the share of reference boundaries in each category, averaged over the rows.}
\label{tab:class}
\begin{tabular*}{\columnwidth}{@{}l@{\hspace{2pt}}lc@{\extracolsep{\fill}}rrrrr|rrrrr@{}}
\toprule
& & \scalebox{0.8}[1]{Grid} & \multicolumn{5}{c}{Buckeye test clean} & \multicolumn{5}{c}{Buckeye test noisy}\\
\cmidrule(lr){4-8}\cmidrule(lr){9-13}
& & \scalebox{0.8}[1]{(ms)} & $\mathrm{M2}_{\scriptscriptstyle I}$ & $\mathrm{M1}_{\scriptscriptstyle I}$ & $\mathrm{M1}_{\scriptscriptstyle B}$ & $\mathrm{M1}_{\scriptscriptstyle E}$ & $\mathrm{M0}$ & $\mathrm{M2}_{\scriptscriptstyle I}$ & $\mathrm{M1}_{\scriptscriptstyle I}$ & $\mathrm{M1}_{\scriptscriptstyle B}$ & $\mathrm{M1}_{\scriptscriptstyle E}$ & $\mathrm{M0}$\\
\midrule
& \emph{\% of boundaries} & & 69.2 & 10.4 & 7.2 & 7.6 & 5.5 & 64.4 & 12.2 & 6.7 & 6.9 & 9.7\\
\midrule
\multirow{8}{*}{\smash{\rotatebox[origin=c]{90}{\tiny Open}}} & Whisper-ts & 20 & .17 & .08 & .58 & .03 & .14 & .17 & .09 & .57 & .02 & .14\\
 & Whisper & 20 & .12 & .06 & .57 & .05 & .13 & .12 & .07 & .57 & .04 & .14\\
 & Parakeet & 80 & .23 & .15 & .22 & .04 & .08 & .23 & .16 & .23 & .03 & .09\\
 & W$\to$WhisperX & 20 & .13 & .07 & .10 & .28 & .06 & .13 & .08 & .08 & .25 & .08\\
 & TorchAudio & 20 & .13 & .13 & .14 & .09 & .15 & .12 & .11 & .11 & .09 & .13\\
 & CrisperW & 10 & .52 & .34 & .56 & .21 & .20 & .47 & .31 & .55 & .22 & \textbf{.22}\\
 & Q$\to$Qwen3-FA & 80 & .43 & .30 & .55 & .35 & .21 & .40 & .26 & .52 & .27 & .17\\
 & Q$\to$MFA & 10 & .73 & .47 & .57 & .33 & .22 & .64 & .36 & .56 & .22 & .18\\
\cmidrule(lr){1-12}
\multirow{11}{*}{\smash{\rotatebox[origin=c]{90}{\tiny API}}} & Deepgram & 80 & .21 & .12 & .48 & .14 & .16 & .21 & .13 & .48 & .14 & .14\\
 & Speechmatics & 40 & .18 & .10 & .16 & .01 & .06 & .17 & .09 & .16 & .01 & .06\\
 & Azure & 10 & .17 & .12 & .09 & .11 & .06 & .16 & .10 & .08 & .11 & .06\\
 & IBM & 20 & .24 & .16 & .14 & .01 & .07 & .22 & .15 & .12 & .02 & .05\\
 & Amazon & 10 & .33 & .29 & \textbf{.62} & .02 & .20 & .32 & .28 & \textbf{.62} & .02 & .18\\
 & AssemblyAI & 1 & .23 & .14 & .39 & .09 & .17 & .22 & .12 & .34 & .08 & .14\\
 & ElevenLabs & 20 & .12 & .10 & .13 & .09 & .09 & .13 & .10 & .14 & .08 & .10\\
 & Google & 40 & .58 & .44 & .13 & .43 & .18 & .58 & .42 & .17 & \textbf{.39} & .17\\
 & Q$\to$Olign & 10 & \textbf{.80} & .50 & .58 & \textbf{.50} & .27 & .70 & .41 & .46 & .31 & .19\\
 & G$\to$Olign & 10 & \textbf{.80} & .53 & .59 & \textbf{.50} & \textbf{.31} & \textbf{.71} & .44 & .47 & .31 & .21\\
 & P$\to$Olign & 10 & \textbf{.80} & \textbf{.57} & .59 & \textbf{.50} & .29 & \textbf{.71} & \textbf{.47} & .46 & .31 & .20\\
\bottomrule
\end{tabular*}
\end{table}

\label{sec:results}

Word and phone tier results are shown in Tab.~\ref{tab:main} and~\ref{tab:phone}. The results of the dev sets and more
details are in \texttt{records/}\recfn.

\textbf{Clean-speech MAE rankings do not predict robustness.}
MAPS~\cite{kelley2024maps} is third of the fourteen
Track~1 aligners on clean TIMIT word MAE and last under degradation, and four
of the fourteen move three or more places on TIMIT and seven on Buckeye. The $F_1$ ranking
barely moves, because a gross error inflates MAE but is only one miss for $F_1$.

\textbf{Phone error rate measures the lexicon.}
All Track~1 systems in Table~\ref{tab:phone} are fed the reference word
sequence, except FALCON, which has no English G2P~\cite{rousso2026falcon} and
is fed the reference phones instead, and thus leads the phone $F_1$ on TIMIT.
For the others, the clean-audio PER of 26 to 36\% measures a dictionary
against a hand transcription. Degrading the audio moves the PER column by 0.3
points or less for six of the eight systems on TIMIT and five on Buckeye while
their phone MAE moves by up to 82\,ms. Since both phones beside a boundary must match, the same mismatch
caps the phone $F_1$, and only 43 to 54\% of the TIMIT phone boundaries have
both neighbours matched.

\textbf{Commercial APIs can suffer from conservativeness.}
Under noisy conditions, a conservative aligner could loosen utterance
boundaries into silence to prevent possibly cutting the edge words, which can
cost its edge boundary precision. Olign places
interior word boundaries better than MFA, 9.3\,ms against
14.3\,ms clean and 15.0\,ms against 19.7\,ms degraded, and overshoots into the
silence at a word end, 6\,ms clean and 25\,ms degraded where MFA overshoots
14\,ms and 8\,ms. That flips its noisy TIMIT MAE to 33.0\,ms against
30.2\,ms. Averaged over clean and degraded, Olign overshoots by 15\,ms and
Google Chirp~2 by 56\,ms.

\textbf{Re-aligning ASR output improves over the bundled aligner.}
Track~2 hands the recognizer's words to the aligner, and its errors come
along but its timestamps need not. Google Chirp~2 followed by
Olign~\cite{olewave2024olign} improved the $F_1$ at 20\,ms by 40\% on average
over the four test cells.

\textbf{ASR accuracy affects FA accuracy.}
Recognition errors directly lower the alignment score by leaving boundaries with only one matched label. These $\mathrm{M1}_{I}$ boundaries from Table~\ref{tab:class} never count as hits in the overall $F_1$ of Tables~\ref{tab:main} and~\ref{tab:phone}. When average Track~2 WER on Buckeye worsens from 12.6\% clean to 17.7\% degraded, this unscorable $\mathrm{M1}_{I}$ group expands from 10.4\% to 12.2\%. Aligners also time boundaries much better when they recognize both adjacent words. The mean $F_1$ across Track~2 systems on clean Buckeye is 0.36 for $\mathrm{M2}_{I}$ boundaries and only 0.25 for $\mathrm{M1}_{I}$. A stronger recognizer fixes both problems by shifting boundaries into $\mathrm{M2}_{I}$ where they count and are timed more accurately.

\textbf{Utterance end timestamps are more difficult to estimate than beginning
and internal ones.}
As shown in Table~\ref{tab:class}, the boundary detection accuracy is related
to the location of the boundary in an utterance. Averaged over the Track~2
systems on Buckeye test, the $F_1$ of utterance end boundaries is 0.18 and
0.17 lower than that of beginning and internal boundaries respectively.

\textbf{Word start and end times carry large systematic bias.}
As shown in Fig.~\ref{fig:bias-timit-core_test-word}, a system's mean signed
error is rarely near zero. Whisper is about 150\,ms early and seven of the
nineteen plotted are more than 50\,ms out at the word start. Absolute error
cannot see this. Whisper sits on the
dashed diagonal, about as early at both ends, so its word durations are right
to within 16\,ms and a duration check would pass timestamps 150\,ms early. Seven of the eight commercial ASR APIs start their
words late, three of them by more than 50\,ms.

\textbf{A coarse timestamp grid caps detection.}
Some systems report times only in 80\,ms steps, the Grid column of
Table~\ref{tab:class}. Only half of the reference boundaries fall within
20\,ms of such a step, so a forced aligner on that grid is statistically
capped near an $F_1$ of 0.5 at 20\,ms. Qwen3-FA reports on 80\,ms and its 0.40
in Table~\ref{tab:main} sits under that cap.

\section{Conclusions}
\label{sec:concl}

FA-Bench is a common evaluation setting for forced alignment and for the
timestamps an ASR system emits. It fixes transcript normalization, held-out
splits and the phone inventory, and releases the scoring code. By running
\mbox{FA-Bench} on human-annotated datasets, we show that word error rate
alone does not determine timestamp accuracy, that a clean-speech MAE ranking does not survive degradation, and that word boundaries carry offsets that absolute
error never sees. Every boundary is scored, utterance edges included, and a
boundary counts only when both labels beside it match the reference. Grouping
boundaries by where they sit and by how many adjacent labels match turns one
number into a statement about where a system lost it. Future releases will add
more languages and systems, and more options for users.

\let\oldthebibliography\thebibliography
\renewcommand{\thebibliography}[1]{%
  \oldthebibliography{#1}%
  \fontsize{9pt}{9.2pt}\selectfont
  \setlength{\itemsep}{0pt}%
  \setlength{\parsep}{0pt}%
  \setlength{\parskip}{3.5pt plus 2pt minus 2.5pt}%
}
\bibliographystyle{IEEEbib}
\bibliography{refs}

\end{document}